\documentclass[review]{elsarticle}
\usepackage[utf8]{inputenc}
\usepackage{multirow}
\usepackage{mdwtab}
\usepackage{graphics} % for pdf, bitmapped graphics files
\usepackage{epsfig}
\usepackage{graphicx}
\usepackage{epsfig} % for postscript graphics files
\usepackage{mathptmx} % assumes new font selection scheme installed
\usepackage{times} % assumes new font selection scheme installed
\usepackage{amsmath} % assumes amsmath package installed
\usepackage{amssymb}  % assumes amssymb package installed
\usepackage[table,xcdraw]{xcolor}

\journal{Neural Networks}

\begin{document}

\begin{frontmatter}

\title{Guiding GANs: How to control non-conditional pre-trained GANs for conditional image generation}

\author{Manel Mateos, Alejandro Gonz\'alez\footnote{Corresponding author: a.gonzalez@salle.url.edu}, Xavier Sevillano}
\address{GTM - Grup de Recerca en Tecnologies M\`edia. La Salle - Universitat Ramon Llull}
%\ead{{manel.mateos,a.gonzalez,xavier.sevillano}@salle.url.edu}

\begin{abstract}
Generative Adversarial Networks (GANs) are an arrange of two neural networks --the generator and the discriminator-- that are jointly trained to generate artificial data, such as images, from random inputs. The quality of these generated images has recently reached such levels that can often lead both machines and humans into mistaking fake for real examples. However, the process performed by the generator of the GAN has some limitations when we want to condition the network to generate images from subcategories of a specific class. Some recent approaches tackle this \textit{conditional generation} by introducing extra information prior to the training process, such as image semantic segmentation or textual descriptions. While successful, these techniques still require defining beforehand the desired subcategories and collecting large labeled image datasets representing them to train the GAN from scratch. In this paper we present a novel and alternative method for guiding generic non-conditional GANs to behave as conditional GANs. Instead of re-training the GAN, our approach adds into the mix an encoder network to generate the high-dimensional random input vectors that are fed to the generator network of a non-conditional GAN to make it generate images from a specific subcategory. In our experiments, when compared to training a conditional GAN from scratch, our guided GAN is able to generate artificial images of perceived quality comparable to that of non-conditional GANs after training the encoder on just a few hundreds of images, which substantially accelerates the process and enables adding new subcategories seamlessly.
\end{abstract}


\end{frontmatter}

\section{Introduction}
The generation of artificial data that follows real distributions has encouraged the computer science community to develop generative algorithms that aim to create data as indistinguishable as possible from real data. Applications range from the generation of missing data for incomplete datasets \cite{Hartley:1958} to coherent text generation \cite{Roh:2003}, among many others.

%such as the generation of basic random vectors that follow a given clustered distribution (mean and variance), or the regression for artificial estimation of an unseen real example, up to coherent text generation or the generation of missing data (data, text, image, ...) that completes an incomplete dataset, are just some examples of what has been tested, and achieved before.\\

Recently, the computer vision community has focused on the generation of real-looking artificial images, and a specific type of neural networks called generative adversarial networks (GANs) have attained remarkable performance in this area \cite{Alqahtani:2019}. In GANs, models  are  built  with  two  neural  networks: the generator, which is a convolutional neural network (CNN) trained to generate images that mimic the distribution of the training dataset, and the discriminator, which tries to distinguish between real images and fake images generated by the generator. The process of the generator trying to fool the discriminator leads to a joint learning process that allows to effectively generate fake real-looking images similar to those in the training set.

Despite their success, giving users total control on the characteristics of the images generated by a GAN is still an issue to be solved. Traditionally, researchers have used additional information as inputs in the generator network training to condition the generation process \cite{Dai:2017}. This gives rise to \textit{conditional GANs}, which are able to generate artificial images with certain specific, desired characteristics. Many authors in recent years have explored different ways of including this conditional information, be it through image textual descriptions \cite{Chang:2019}, semantic segmentations \cite{Tang:2020a} or image category definition \cite{Qi:2018}, among others. However, all these approaches set the number and definition of characteristics \textit{before} the GAN training process and cannot be changed later. Thus, for any addition or variation, a new labeled dataset must be collected, and the GAN must be re-trained, making the whole process complex and time-consuming. 

\begin{figure}[t]
\centering
\includegraphics[width=0.6\linewidth]{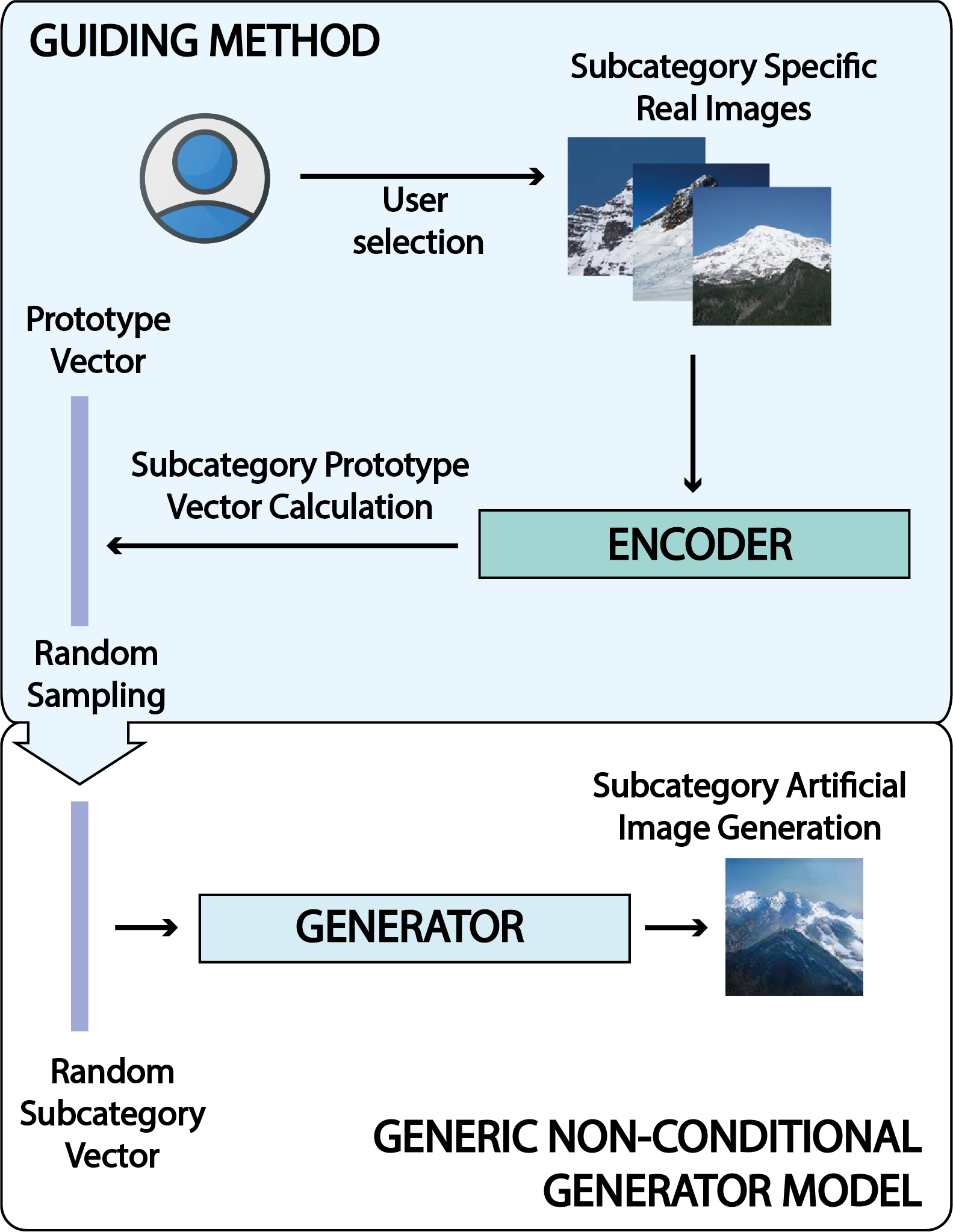}
\caption[Guiding GANs.]{Guiding GANs process: the user collects a small sample of images corresponding to the specific subcategory. Based on this sample, the encoder network generates the subcategory prototype vector that represents the distribution of the sample. This vector is then randomly sampled to create random vectors that are fed to the pre-trained non-conditional GAN to generate images of the desired subcategory.}
\label{G-GANs:fig:Graphical abstract}
\end{figure}

In the face of these limitations of current conditional GANs, in this paper we present a novel and efficient way of guiding pre-trained non-conditional GANs for generating images belonging to specific subcategories avoiding burdensome re-training processes. In Figure \ref{G-GANs:fig:Graphical abstract}, the general scheme of the proposed method is depicted. 

The key element of our proposal is an encoder network, which is first trained to learn the opposite transformation of the pre-trained non-conditional GAN. To do so, pairs of random vectors and the corresponding images generated by the GAN are employed.

Then, the guiding process starts with the user collecting a small sample of non-annotated real images as examples of the specific image subcategory he/she wants the pre-trained non-conditional GAN to generate. These images are fed to the encoder network, which estimates the distribution of the selected subcategory, and embeds it in the so-called subcategory prototype vector. By randomly sampling the prototype vector, we create multiple random vectors which are fed to the pre-trained generator network of the GAN to produce artificial images similar to the previously selected ones. By proceeding this way, we enable users to obtain real-looking artificial images at a reduced computational cost. 

This paper is organized as follows: in Section \ref{sec:SotA} the theoretical background of our proposal, as well as relevant previous work is reviewed. Then, Section \ref{sec:Method} describes the proposed method. Next, in Section \ref{sec:ExpRes}, our proposal is evaluated in a series of experiments. Finally, Section \ref{sec:Con} discusses the obtained results, highlights the advantages of our method and outlines future research lines.

\section{State of the art}
\label{sec:SotA}
The image generation problem has pushed researchers to find the most optimal frameworks and models to produce real-looking artificial images. 

In addition to GANs \cite{Goodfellow:2014}, other relevant approaches include adversarial auto-encoders (AAEs) \cite{Makhzani:2015}, variational auto-encoders (VAEs) \cite{Kingma:2013} and auto-regression models (ARMs) (e.g. PixelRNN \cite{oord:2016}). 

As GANs are probably the most widely employed technique for the conditional generation of high resolution artificial images, this section is focused on reviewing the basic concepts of GANs (Section \ref{sec:GANs}), as well as common approaches to GAN-based increased resolution image generation (Section \ref{sec:progressiveGANs}) and proposals on conditional generation image generation via GANs (Section \ref{sec:conditionalGANs}).

\subsection{Artificial image generation via GANs}
\label{sec:GANs}

\begin{figure}[ht]
\centering
\includegraphics[width=0.9\textwidth]{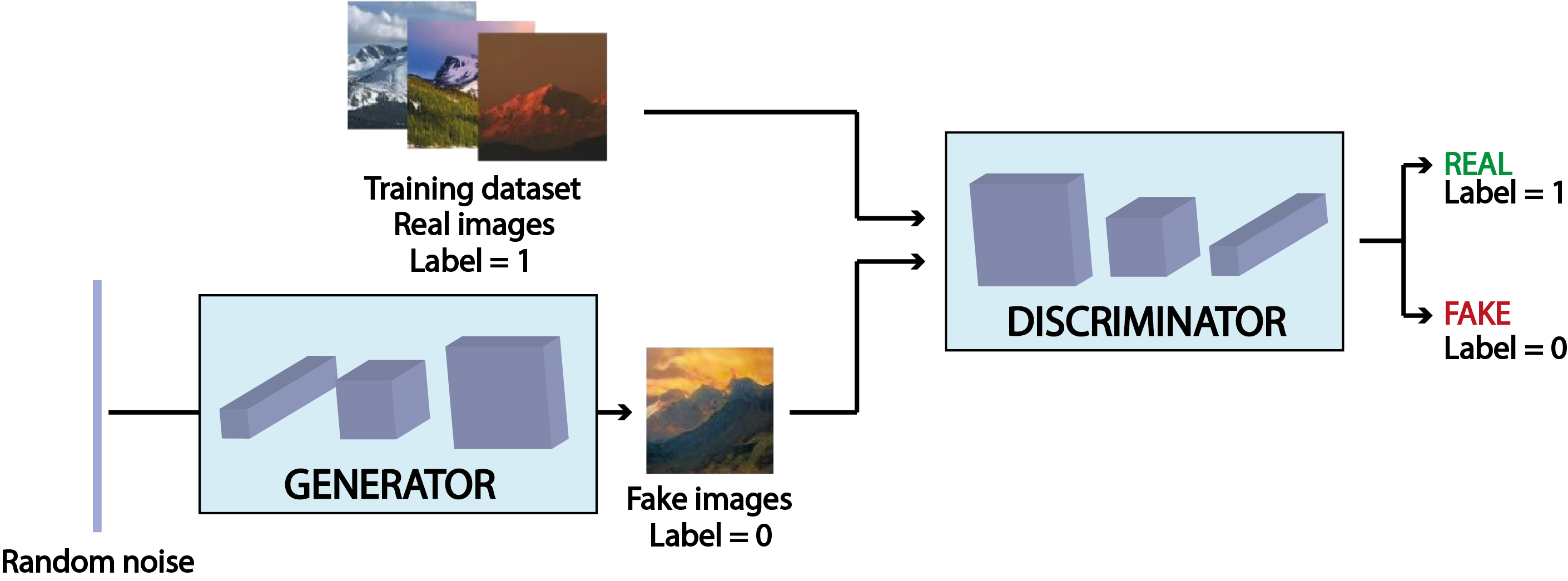}
\caption[Generic GAN architecture.]{Generic GAN architecture. Generator, discriminator and their connections.}
\label{G-GANs:fig:GANS_Basics}
\end{figure}

GANs are a subset of implicit density generative models that focus on transforming a random input into an image which aims to be part of the true distribution of the training data set \cite{Goodfellow:2014}. As mentioned earlier, the GAN models are built through the interaction of two neural networks, the generator and the discriminator, that together learn how to generate new real-looking images effectively by trying to confuse one another. On the one hand, the generator network (G) is a CNN that takes a random vector from a distribution and maps it through the model to an output with the desired sample size. Its objective is to produce images that look as similar as possible to the training examples. On the other hand, the discriminator (D) is a binary classification CNN that predicts if a given image is a real one (i.e. a training sample) or a fake one (generated by G). In this sense, the GAN training process can be understood as a two-player game where G tries to fool D by generating real-looking images and D tries to distinguish between real and fake images, as shown in Figure \ref{G-GANs:fig:GANS_Basics}.

These two models are trained jointly minimizing the possible loss for a worst-case scenario by using as the objective function: $ \min\limits_{G}\max\limits_{D}V(D,G) = E_{x\sim p_{data}(x)}[log D(x)] + E_{z\sim p_z(z)}[log(1-D(G(z)))]$, where $D(x)$ is the discriminator output for the real data sample $x$, and $D(G(z))$ is the discriminator output for the artificial data generated by the generator with $z$ as the random input vector.

During the training process, the discriminator tries to maximize the objective function by making $D(x)$ close to 1 and $D(G(z))$ close to 0, while the generator focuses on minimizing it such that $D(G(z))$ is close to 1, fooling the discriminator.

\subsection{Increasing artificial image resolution via progressive growing GANs}
\label{sec:progressiveGANs}

The generation of high resolution and complex images via GANs requires up-scaling their architecture. In this sense, higher resolutions imply some additional challenges, such as \textit{i)} gradients during training become useless, generating poor images easily identifiable as fake, and therefore making the training process fail, \textit{ii)} memory constraints increase, forcing researchers to reduce batch sizes, which compromises the stability of the training process, and \textit{iii)} better hardware acceleration is needed to train these bigger models and handle them efficiently. 

\begin{figure}[ht]
\centering
\includegraphics[width=0.8\textwidth]{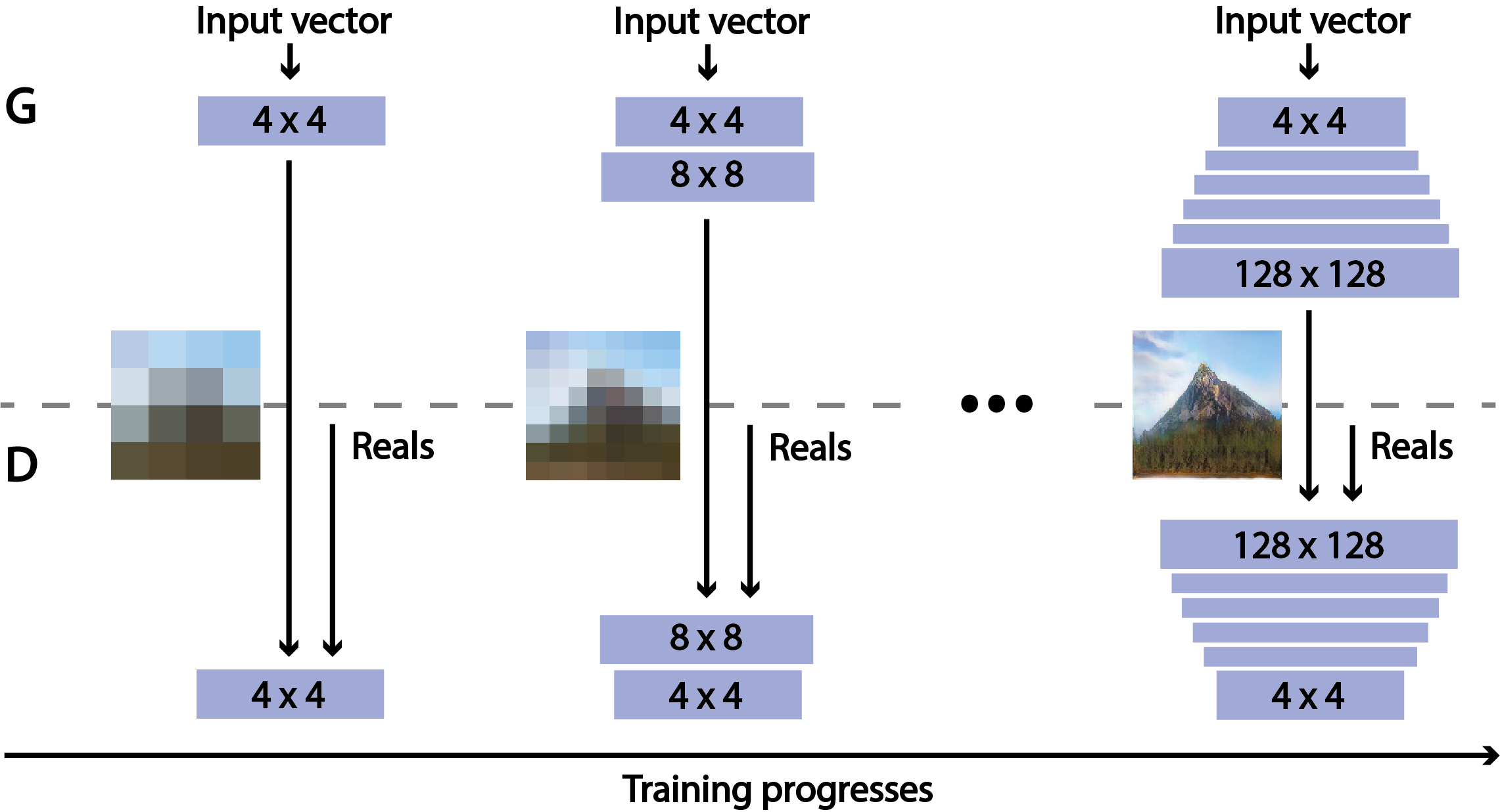}
\caption[Progressive GANs training process.]{Progressive GANs training process. The training process for G and D starts at a low-resolution ($4 \times 4$ pixels) and is gradually adapted to higher resolutions by adding layers to G and D. At each training step all existing layers remain trainable. Added layers are convolutional layers operating on $N \times N$ spatial resolution.}
\label{G-GANs:fig:Prog_GANS_Train}
\end{figure}

These problems forced the research community to formulate new approaches to scale up the resolution of the generated images successfully. Some authors like Salimans et al. \cite{Salimans:2016} and Gulrajani et al. \cite{Gulrajani:2017} presented improvements on the GAN training process, and others like Berthelot et al. \cite{Berthelot:2017} and Kodali et al. \cite{Kodali:2017} proposed new GAN architectures.

An alternative and interesting approach are Progressive Growing GANs (PG-GANs) \cite{Karras:2017}, which consist in gradually training GANs and iteratively adapting them to higher resolution images on each step of the training. The authors propose starting the training process on low-resolution images and then gradually increasing the resolution by adding layers to both G and D (see Figure \ref{G-GANs:fig:Prog_GANS_Train}). This procedure implies that G and D must be symmetrical and grow synchronously. The method is based on the premise that CNNs are capable of first learning large general features, generalizing the training images, and the addition of more layers allows to move into the finer details. According to Karras et al., the use of this progressive growing approach \textit{i)} reduces GANs training time, \textit{ii)} improves convergence, as low-resolution neural networks are stable and the progressive increase of image resolution allows starting the higher resolution training process with a stable pre-trained network, and \textit{iii)} introduces an adaptation degree in terms of resolution, allowing the control of the training process for obtaining artificial images of a given desired resolution. Due to this flexibility and reduced training time, we adopt PG-GANs in our work to generate artificial images.

\subsection{Conditional image generation with GANs}
\label{sec:conditionalGANs}

Another issue worth considering is how to control the generation process beyond the training dataset, a process that goes by the name of conditional generation. 

In this context, authors have developed different approaches based on training the GANs not only with uncategorized images but also introducing categorical information of the images included in the training dataset. For instance, works like \cite{Chang:2019} used the categories of the training images as extra features in the generation and discrimination processes. 

Other authors addressed the problem by using both categorical information of the training images and also their semantic segmentation, as in \cite{Tang:2020a,Qi:2018,Bau:2019,Park:2019}. In those works, the authors train the discriminator to distinguish real and fake images and at the same time, to match the objective pixel wise semantic information given.

An example of a use case for conditional GANs consists on generating human body images simulating specific body poses. In this context, some authors proposed new architectures for the training of GANs, which receive the body poses of the training images as an extra feature. \cite{Tang:2019,Dong:2018,Ma:2018}, once the GAN is trained users may generate images where the pose is freely chosen.

The following section describes our proposal, which guides non-conditional PG-GANs for generating images of a specific subcategory within the training set, at will and without the need of retraining the GAN models. 

\section{Guiding non-conditional pre-trained GANs}
\label{sec:Method}

Our proposal is based on considering the $d$-dimensional input space of random vectors that feed the generator network G of a non-conditional GAN once it is trained to generate images of a specific category $C$, which we refer to as $\mathrm{G}_C$. Without loss of generality, any category $C$ is intrinsically composed of multiple (say $m$) subcategories $SC_i$, that is $C=\displaystyle\bigcup_{i=1}^{m}{SC_i}$.

In response to these random input vectors, $\mathrm{G}_C$ generates images corresponding to the category represented in the training set, but no control mechanism is available to ``tell'' $\mathrm{G}_C$ to generate images of a specific subcategory $SC_k$. In our method, we propose hand-picking the random vector input to $\mathrm{G}_C$ to produce images belonging to the desired subcategory, thus giving the user total control over the artificial image generation process. 

Our method is described step by step in the following paragraphs. Please refer to Figure \ref{G-GANs:fig:Graphical abstract} for a graphical reference.

\begin{figure}[t]
\centering
\includegraphics[width=0.75\textwidth]{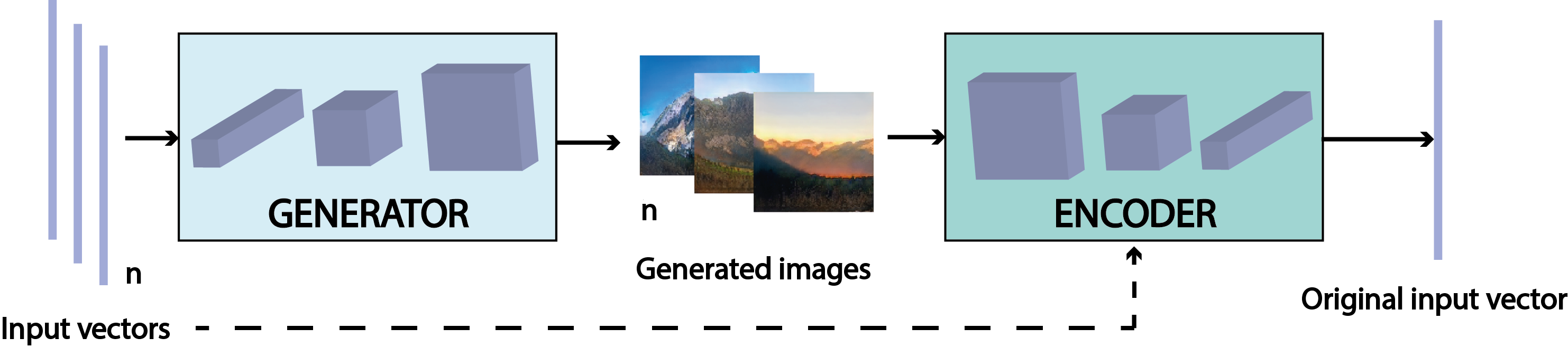}
\caption[Encoder training.]{Encoder training.}
\label{G-GANs:fig:GAN_Encoder}
\end{figure}

{\bf Step 1) Encoder training:} an encoder network is trained to learn the opposite transformation from the one carried out by the trained generator $\mathrm{G}_C$. To that end, the encoder is trained using pairs of $d$-dimensional random input vectors and their correspondent generated images directly extracted from the non-conditional generator network $\mathrm{G}_C$, therefore making the training data supply virtually endless (see Figure \ref{G-GANs:fig:GAN_Encoder}). As a result, we obtain an encoder model that given an image generated by $\mathrm{G}_C$ returns the $d$-dimensional input vector which would have created that image. After finishing the training process, the encoder is capable of successfully returning input vectors from random images not produced by the generator.% That is, feeding the input vector returned by the encoder into the generator results in similar images, even for examples never produced before by the generator.\\

{\bf Step 2) Subcategory random vectors generation:} The user collects real images of the desired subcategory $SC_k$ and feeds the trained encoder with them. This image collection process can be fully automated, as described in Section \ref{sec:ExpRes}. In response, the encoder returns a $d$-dimensional random vector $\vec{x}_{i}^{SC_k}$ corresponding to each input image. Notice that the larger the number of collected images (referred to as $N$), the more accurate the estimation of the distribution of the desired subcategory. Moreover, it is also to note that the user can decide to add a new subcategory, and the obtainment of the corresponding vectors through the encoder can be started at any point.% Around 500 images are often sufficient to achieve good results on any new class.

{\bf Step 3) Subcategory prototype vector creation and sampling:} the mean value and standard deviation of each of the $d$ components of the vectors $\vec{x}_{i}^{SC_k}$ ($\forall i=1...N$) output by the encoder in response to the images of the desired subcategory $SC_k$ are computed and embedded in the subcategory prototype vector $\vec{p}^{SC_k}$. Next, this prototype vector is used to generate as many random vectors as desired by sampling $d$ normal random variables $X_j \sim N\left(\mu_j,\alpha\cdot\sigma_j \right)$ (with $\forall j=1..d$), where $\mu_j$ and $\sigma_j$ are the mean value and the standard deviation of the $j$th component of the vectors $\vec{x}_{i}^{SC_k}$ ($\forall i=1...N$), and $\alpha$ is a scalar parameter. In our experiments, we heuristically tuned the value of this parameter to 2.5. These random vectors follow the distribution of the desired subcategory $SC_k$, so they will make the pre-trained generator network $\mathrm{G}_C$ generate images belonging to the specific subcategory of choice.

%To do so, the mean and standard deviation of each component of all vectors $\vec{x}_{i}^{SC_k}$ is computed  AQUI Lastly and per class, all the input vectors stored can be processed to create a master vector. This vector is the base for generating new guided images. This master vector results from computing the median per every element inside all input vectors for an specific class, and then using this information together with the standard deviation per position and sample to randomly create new input vectors which produced random images that belong to the new guided class.\\

\begin{figure}[h]
\centering
\includegraphics[width=0.8\linewidth]{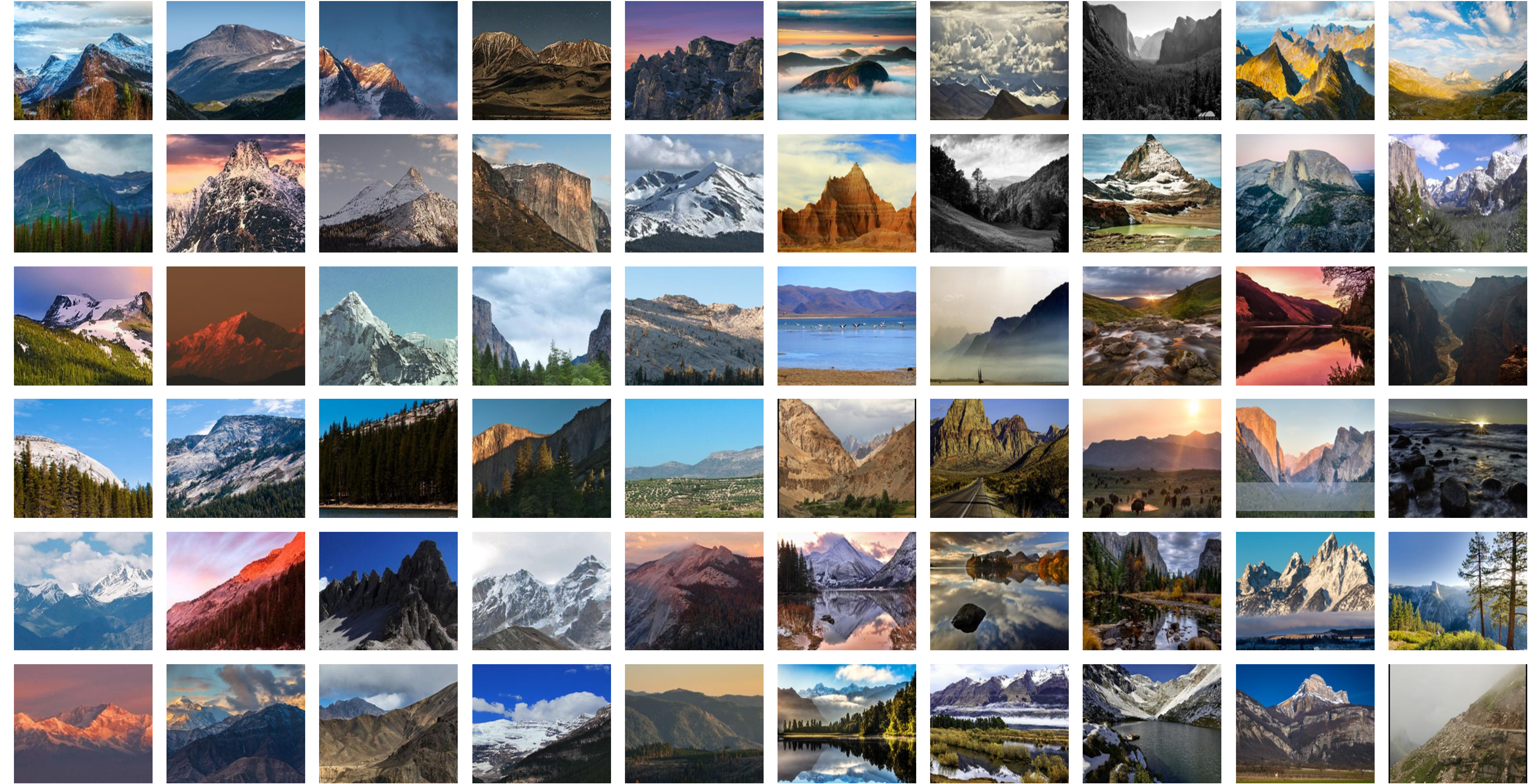}
\caption[Generic GAN]{Training real examples of the ``mountains'' category used to train the non-conditional progressive growing GAN}
\label{G-GANs:fig:Generic_GAN_Training_Dataset}
\end{figure}

\section{Experiments and results}
\label{sec:ExpRes}
The experiments described in this section aim to evaluate our method to guide a non-conditional progressive growing GAN. 

We start by describing the data employed in our experiments. Subsequently, we present the architecture of the PG-GAN, and an experiment involving a subjective quality evaluation test to assess the quality of the images it generates.

In the final experiment, we guide the non-conditional PG-GAN to generate images from specific subcategories of choice. We describe the architecture of the encoder network employed in the experiments, and then evaluate \textit{i)} the ability of the non-conditional network to effectively generate images that correspond to the chosen subcategories, and \textit{ii)} the perceived quality of the generated images. 

\subsection{Dataset}
\label{sec:dataset}

The non-conditional PG-GAN was trained to generate images of the category $C = $ ``mountains''. To that end, a total of 19.765 images of mountains were downloaded from the Flickr image hosting service and used to train the GAN. Some example images from the training dataset are presented in Figure \ref{G-GANs:fig:Generic_GAN_Training_Dataset}. 

On the other hand, to train the encoder network, we created 500.000 random vectors, fed them to the pre-trained non-generic PG-GAN, and collected the corresponding images.  

The images used for guiding the non-conditional GAN to generate images from a specific subcategory were obtained using the Flickr API, downloading $N$ images that were tagged as one of the following selected subcategories $SC_k = \{$``mountains + snow'', ``mountains + sunset'', ``mountains + trees'', ``mountains + night'' and ``mountains + rocks''$\}$. 

\subsection{Non-conditional progressive growing GAN}
\subsubsection{Architecture}

The PG-GAN used in these experiments follows the architecture presented by Karras et al. in \cite{Karras:2017} and shown in Figure \ref{G-GANs:fig:Prog_GANS_Train}. 

In a nutshell, the model starts training at a resolution of 4x4 and progresses until reaching a final resolution of 128x128. 

The architecture of both the generator and the discriminator are based on strided convolutions with leakyReLU activations and constrain the signal magnitude and competition during training through pixel wise feature normalization and equalizing the learning. The whole model has over 45 million parameters and was trained on Google Colab for 200 epochs.

\begin{figure}[t]
\centering
\includegraphics[width=0.8\textwidth]{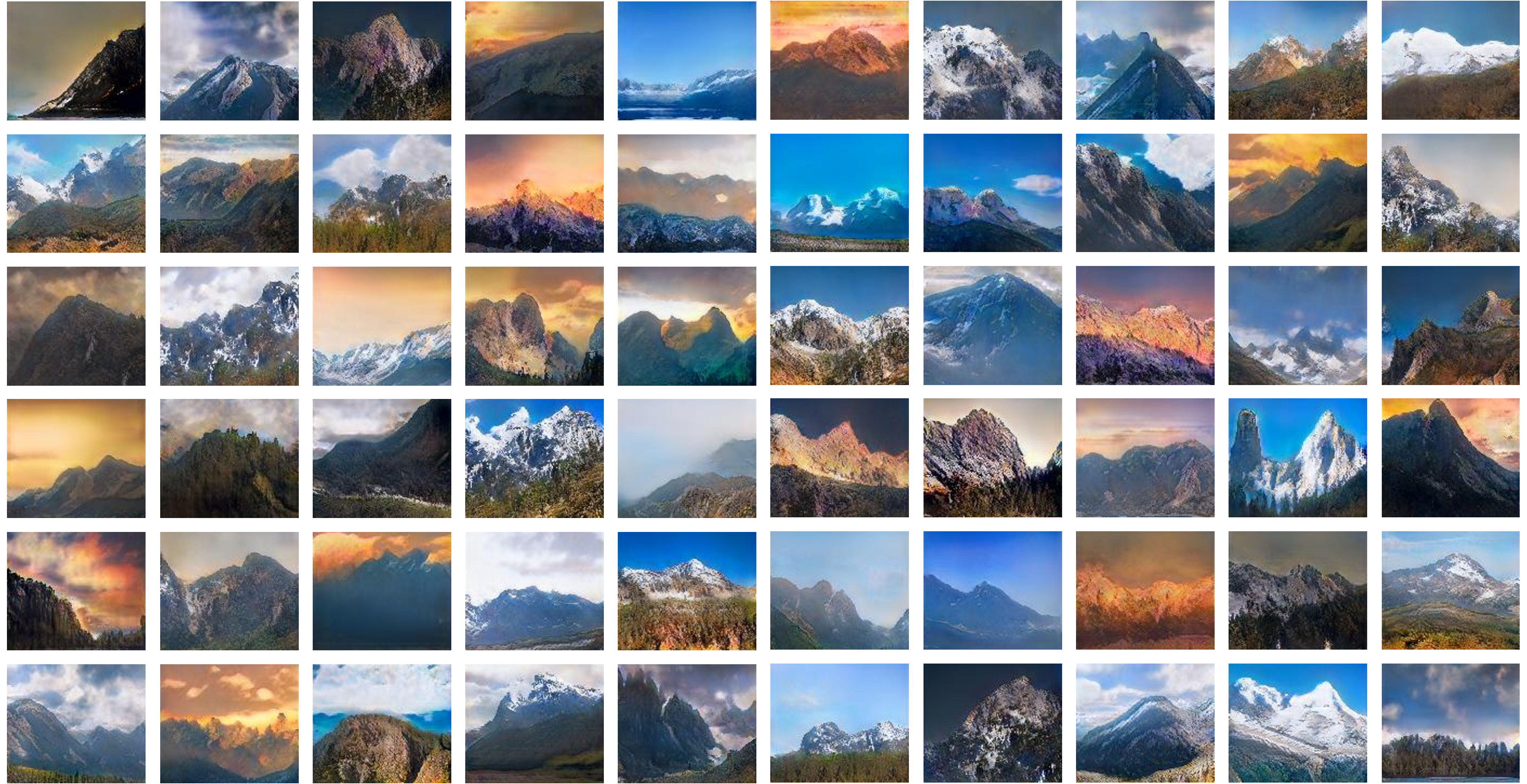}
\caption[Generic GAN]{Examples of the ``mountain'' category artificial images generated by the non-conditional progressive GAN}
\label{G-GANs:fig:Generic_GAN_Samples}
\end{figure}

Figure \ref{G-GANs:fig:Generic_GAN_Samples} shows several examples of the artificial images of the mountain category generated by the GAN. All examples portray high fidelity and variance, successfully capturing the true distribution of the images provided during the training of the model. 

\begin{figure}[h]
\centering
\includegraphics[width=0.75\linewidth]{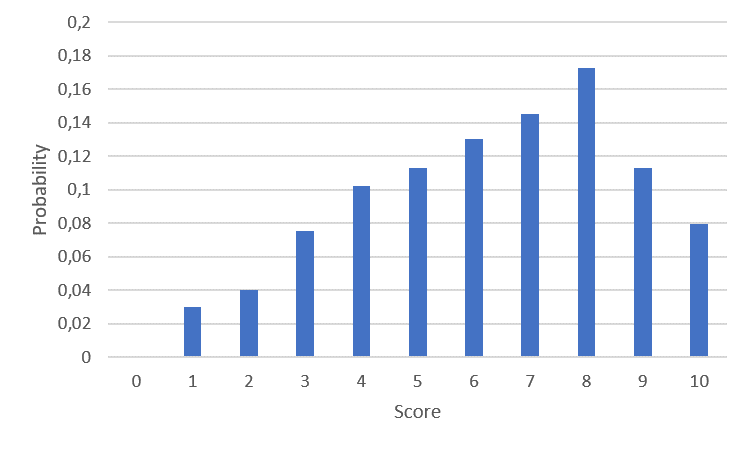}
\caption[Generic GAN]{Normalized histogram of the scores given by the participants in the subjective quality evaluation test of the non-conditional progressive GAN}
\label{G-GANs:fig:Generic_GAN_Perception_Test}
\end{figure}

\subsubsection{Artificial image quality evaluation}

To evaluate the quality of the images generated by the non-conditional PG-GAN, we carried out a subjective quality evaluation test, in which 50 participants were asked to evaluate the degree of realism of 20 artificial images using a rating scale from 0 to 10 (the greater the score, the greater the realism). The normalized histogram of the obtained ratings is depicted in Figure \ref{G-GANs:fig:Generic_GAN_Perception_Test}. The left skewed distribution of scores reveals that the participants judged most of the images as quite realistic, obtaining an mean opinion score of 6.3.

\subsection{Guiding the non-conditional GAN}
\begin{figure}[hb]
\centering
\includegraphics[width=0.85\linewidth]{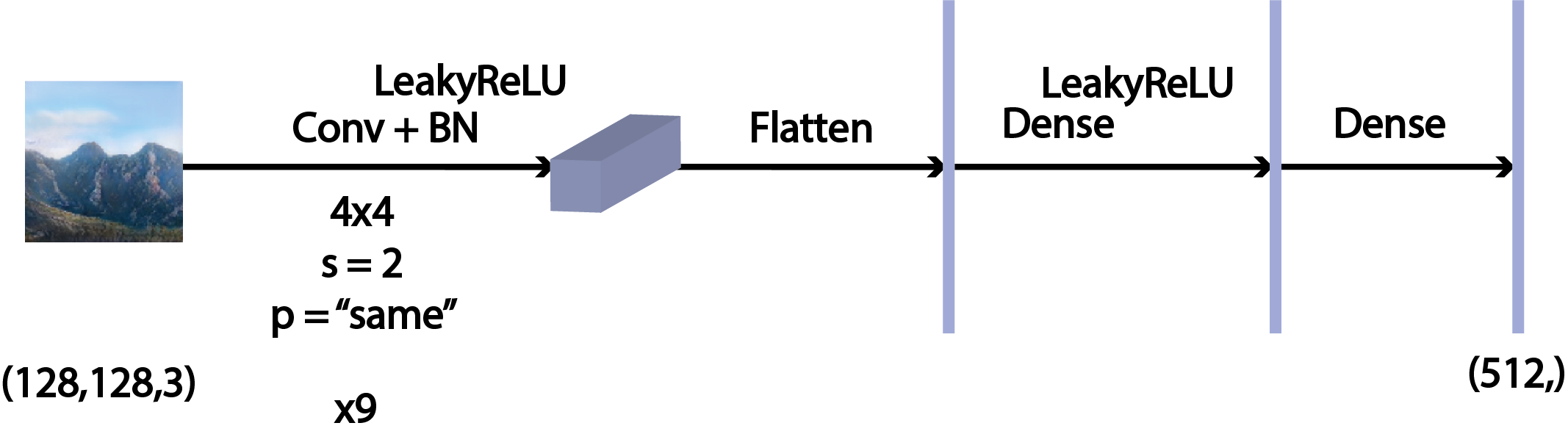}
\caption[Encoder]{Architecture of the encoder network}
\label{G-GANs:fig:Encoder_Architecture}
\end{figure}

The architecture of the encoder network used to guide the non-conditional GAN is presented in Figure \ref{G-GANs:fig:Encoder_Architecture}. 

As mentioned earlier, the encoder was trained on 500.000 pairs of random vectors and the corresponding artificial images generated by the non-conditional PG-GAN. The training took 4 epochs to converge.

The computation of the subcategory prototype vector was made after programmatically  downloading a variable number $N$ of images from Flickr corresponding to the desired subcategories that were described in section \ref{sec:dataset}. 

The experiments to evaluate the quality of the images generated by the guided non-conditional PG-GAN are presented next.

\subsubsection{Effect of $N$ on the perceived quality of the images}

First, we evaluated how the number of images fed to the encoder to create the subcategory prototype vector affects the quality of the images that are subsequently generated by the guided non-conditional PG-GAN.

To that end, we presented 50 participants with images generated when the subcategory prototype vector was computed after feeding the encoder network with $N=\{64, 128 \, \mathrm{and} \, 256\}$ images.

\begin{figure}[t]
\centering
\includegraphics[width=0.7\linewidth]{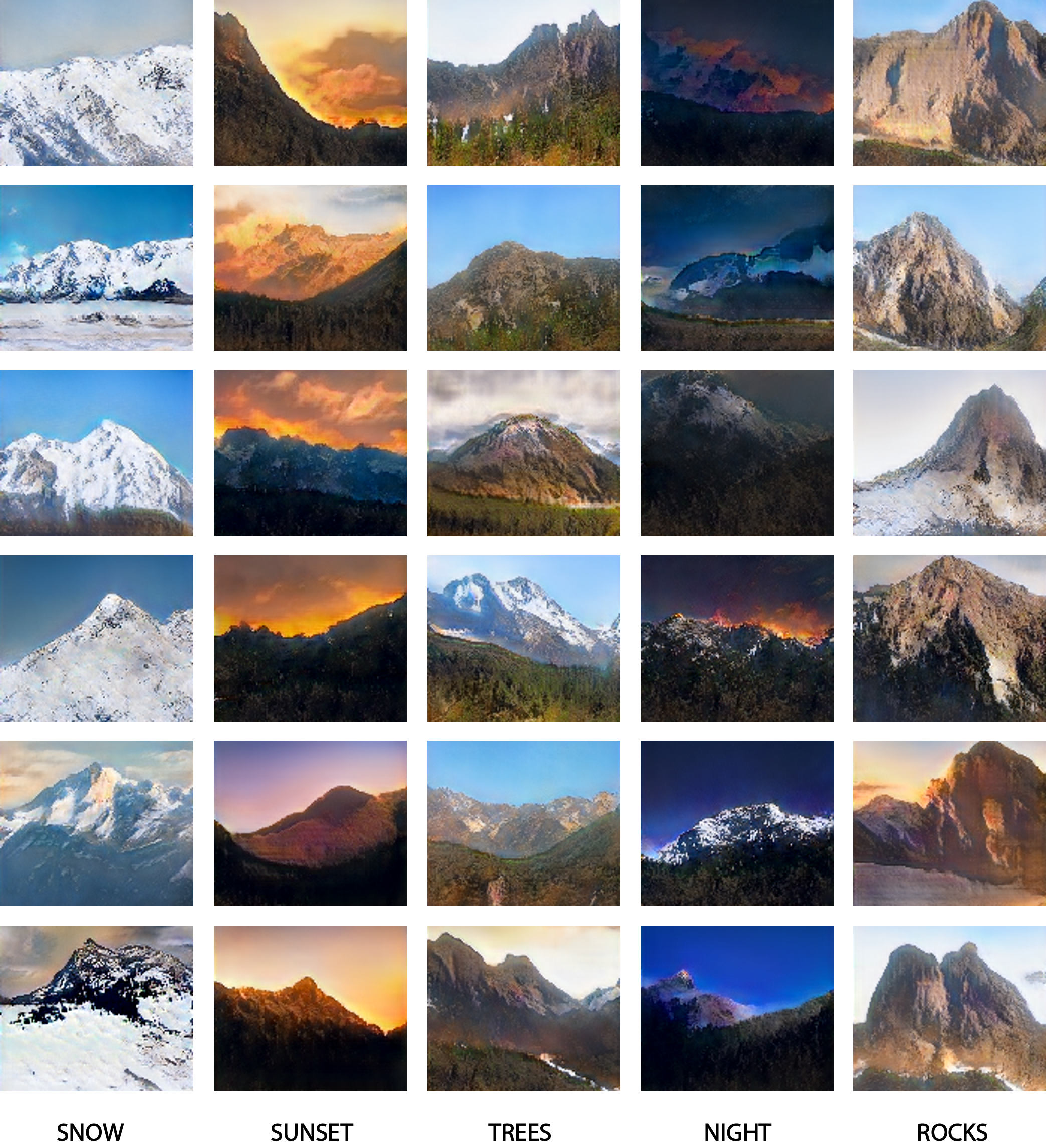}
\caption[Guided GAN]{Examples of images of the ``mountains+snow'', ``mountains+sunset'', ``mountains+trees'', ``mountains+night'' and ``mountains+rocks'' subcategories generated by the guided non-conditional ``mountain'' GAN}
\label{G-GANs:fig:Guided_GAN_Samples}
\end{figure}

The mean opinion score for these configurations was 5.9, 6.2 and 6.4, respectively. Taking into account that the subjective quality evaluation of the images created by the non-conditional progressive growing GAN yielded a mean opinion score of 6.3, these results prove that using a few hundreds of images corresponding to the desired subcategory suffices to generate images of that subcategory with an equivalent level of perceived quality.

To illustrate this fact, Figure \ref{G-GANs:fig:Guided_GAN_Samples} presents images generated by the guided GAN when asked to create images of the subcategories mentioned earlier with $N=256$. It can be observed that the network succeeds in generating images of the specific subcategory.

\subsubsection{Subcategory identification}
In this experiment, we evaluate whether the participants in the subjective evaluation test were able to correctly identify the subcategory of the images generated by the guided GAN.

The experiment consisted of presenting the participants with 20 images that had to be classified in the (``mountains+'') ``snow'', ``sunset'', ``trees'', ``night'' or ``rocks'' subcategories.

In average, the participants successfully chose the correct subcategory with a 85.2\% accuracy. The confusion matrix corresponding to this experiment is presented in Table \ref{G-GANs:fig:Guided_GAN_Confusion_Matrix}. Notice that the ``snow'' and ``sunset'' subcategories are identified close to perfection, while the ``tree'' subcategory is identified with a 56.6\% accuracy. 

\begin{table}[ht]
\centering
\begin{tabular}{cc|c|c|c|c|c|}
\cline{3-7}
& & \multicolumn{5}{c|}{\textit{\textbf{Predicted class}}} \\
\cline{3-7} 
& & \textbf{Snow} & \textbf{Sunset} & \textbf{Trees} & \textbf{Rocks} & \textbf{Night} \\ 
\hline
\multicolumn{1}{|c|}{\multirow{5}{*}{\textit{\textbf{Actual class}}}} & \textbf{Snow} & 99.5 & 0 & 0 & 0 & 0.5 \\ \cline{2-7} 
\multicolumn{1}{|c|}{} & \textbf{Sunset} & 0 & 99.5 & 0.5 & 0 & 0             \\ 
\cline{2-7} 
\multicolumn{1}{|c|}{} & \textbf{Trees}  & 1 & 0.5 & 56.5 & 42 & 0              \\ 
\cline{2-7} 
\multicolumn{1}{|c|}{} & \textbf{Rocks}  & 2 & 0.5 & 7 & 91 & 0              \\ \cline{2-7} 
\multicolumn{1}{|c|}{} & \textbf{Night}  & 1 & 8 & 9 & 2.5 & 79.5
\\ 
\hline
\end{tabular}
\caption[Guided GAN]{Confusion matrix of the subcategory identification experiment (values in \%).}
\label{G-GANs:fig:Guided_GAN_Confusion_Matrix}
\end{table}

\section{Conclusions}
\label{sec:Con}

This work has introduced a novel method that gives users control over the specific type of images generated by GANs. Our proposal enables the generation of artificial images from a user-defined subcategory, guiding a non-conditional GAN thanks to a new architecture that includes an encoder network to feed the GAN.

This novel process transforms the conditional image generation problem into a simpler task for general users, reaching a flexibility level that cannot be reached by a non-conditional GAN. 

Our proposal allows to considerably reduce the time needed to perform conditional image generation, while maintaining similar results in terms of artificial image quality. Additionally, since only a small set of images of the desired subcategory is needed to guide the GAN, the process can be fully automated. Moreover, the proposed method enables the user to select the desired image subcategory on the go, which allows new ideas to be tested in minutes, much faster than the time that would be required to train a new regular, non-generic GAN from scratch.

Moving forward, we believe the subcategory prototype vector creation process described in Section \ref{sec:Method} could be further improved to better represent the subcategory's distribution, which would help the generator network yield more variance between the images belonging to a single subcategory. Additionally, studying how input vectors are transformed throughout the generator process, and specifically trying to understand how dependent the perceived subcategory is to each step of the network, could help better guide the model by not only feeding it the right vector, but also further "steering" the generation process into the desired direction.

\bibliographystyle{elsarticle-num-names}
\bibliography{bib-GANs.bib}

\end{document}